\PassOptionsToPackage{table,xcdraw}{xcolor}
\documentclass[wcp,twocolumn,10pt]{jmlr}

\makeatletter
\renewcommand*{\@jmlrproceedings}{}
\renewcommand*{\@jmlrabbrvproceedings}{}
\def\ps@jmlrtps{%
  \let\@mkboth\@gobbletwo
  \def\@oddhead{}%
  \let\@evenhead\@oddhead
  \def\@oddfoot{\@titlefoot}%
  \let\@evenfoot\@oddfoot
}
\makeatother

\usepackage{booktabs}
\usepackage{makecell}
\usepackage{multirow}
\usepackage{threeparttable}
\usepackage{hyphenat}
\usepackage{enumitem}
\setlist{itemsep=2pt,topsep=4pt,parsep=0pt}
\usepackage{cleveref}

\title[MultiMedVision]{MultiMedVision: Multi-Modal Medical Vision Framework}

\author{%
  \Name{Frank Li} \Email{frank.li@emory.edu}\\
  \addr Emory University
  \AND
  \Name{Bardia Khosravi} \Email{bardia.khosravi@yale.edu}\\
  \addr Yale University
  \AND
  \Name{Mohammadreza Chavoshi} \Email{mohammadreza.chavoshi@emory.edu}\\
  \addr Emory University
  \AND
  \Name{Young Seok Jeon} \Email{youngseokjeon74@gmail.com}\\
  \addr Emory University
  \AND
  \Name{Theo Dapamede} \Email{theo.dapamede@emory.edu}\\
  \addr Emory University
  \AND
  \Name{Hari Trivedi} \Email{hari.trivedi@emory.edu}\\
  \addr Emory University
  \AND
  \Name{Janice Newsome} \Email{janice.newsome@emory.edu}\\
  \addr Emory University
  \AND
  \Name{Judy Gichoya} \Email{judywawira@emory.edu}\\
  \addr Emory University
}

\begin{document}

\maketitle

\begin{abstract}
Multi-modal medical imaging enables comprehensive diagnostics, yet current foundation models process 2D (e.g.\ X-ray) and 3D (e.g.\ CT) data with separate, dimensionality-specific architectures. We present MultiMedVision, a unified framework for joint 2D/3D representation learning built on a Sparse Vision Transformer. Our model uses 3D Rotary Positional Embeddings and variable-length sequence packing to process mixed-modality batches natively within a shared latent space---without modality-specific adapters or treating 3D volumes as 2D slice sequences. Trained with a self-supervised objective on chest X-rays (MIMIC-CXR) and CT scans (CT-RATE), and using a single shared encoder with 5x less data, MultiMedVision achieves competitive performance on both 2D benchmarks (Macro AUROC 0.82 on MIMIC, 0.84 on CheXpert) and 3D tasks (0.85 on CT-RATE). Analysis of the learned representations reveals coexisting modality-specific and shared feature subspaces, demonstrating that unified cross-dimensional representation learning is feasible without sacrificing modality-specific performance.
\end{abstract}

\begin{keywords}
Medical Image Foundation Models, Multi-Dimensional Representation Learning, Self-Supervised Learning
\end{keywords}


\section{Introduction}

\begin{figure*}[t]
    \centering
    \includegraphics[width=0.85\linewidth]{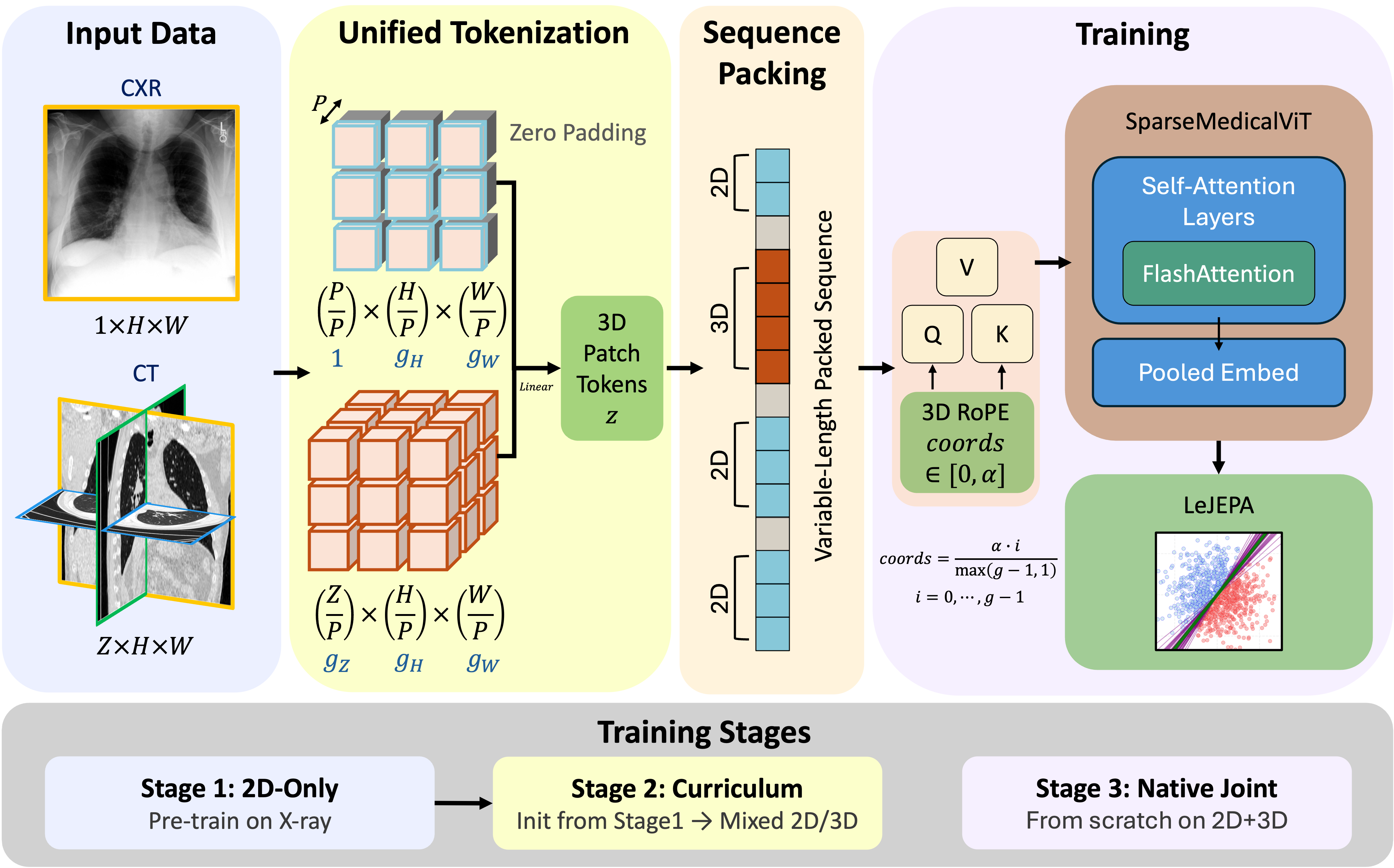}
    \caption{MultiMedVision Overview. 2D and 3D inputs are tokenized into a unified 3D coordinate space via 3D RoPE, then packed into variable-length sequences for efficient processing through a shared SparseMedicalViT encoder. Representations are learned via the LeJEPA self-supervised objective. Three training stages are explored: 2D-only pre-training (Stage1), curriculum adaptation (Stage2), and native joint training (Stage3).}
    \label{fig:Fig1}
\end{figure*}

Medical imaging relies on complementary modalities---2D radiographs (e.g.\ X-rays) for rapid screening and 3D volumetric scans (e.g.\ CT) for detailed structural information. Despite this clinical synergy, current foundation models treat these modalities in isolation. 2D models such as BiomedCLIP \citep{Zhang2025}, MedSigLIP \citep{Sellergren2025}, and RAD-DINO \citep{Prez-Garca2025} cannot process 3D volumes natively, while 3D models like CT-CLIP \citep{Hamamci2026}, 3DINO \citep{Xu2025}, and SPECTRE \citep{Claessens2025} rely on dimensionality-specific architectures or process 3D scans as sequences of independent 2D slices \citep{Liu2025, Mller-Franzes2025}, failing to capture inter-slice spatial continuity. This segregation produces disjoint latent spaces where a patient's X-ray and CT features reside in different manifolds, hindering longitudinal cross-modality analysis. Native integration of 2D and 3D data into a single encoder remains rare: existing unified models such as RadFM \citep{Wu2025} enforce strict modality separation at the batch level, with unification occurring only at the language model stage.

To address these challenges, we present MultiMedVision, a unified framework designed for native 2D/3D representation learning (\cref{fig:Fig1}). Our approach introduces a single, shared training pipeline that can process both 2D and 3D inputs without modality-specific encoders. At its core, MultiMedVision utilizes a Sparse Vision Transformer (SparseMedicalViT) backbone, which builds upon recent advances in flexible tokenization (e.g.\ AToken \citep{Lu2025}, FiT \citep{Lu2024}). Here, \textit{sparse} reflects the fact that the unified 3D coordinate space is only partially occupied depending on the input: a 2D image activates a single planar slice, while a 3D scan populates a full volume. Our architecture employs 3D Rotary Positional Embeddings (RoPE) \citep{Su2024} and variable-length sequence packing to efficiently handle the heterogeneity of medical data. By treating both 2D images and 3D scans as sequences of active tokens within a unified sparse coordinate system, our model learns a unified representation in a shared latent space. We demonstrate the efficacy of this framework using the LeJEPA (Latent Embedding Joint Embedding Predictive Architecture) self-supervised objective \citep{Balestriero2025}, though the framework is flexible enough to support various learning strategies. Our contributions are as follows:
\begin{enumerate}
    \item \textbf{Flexible Training Framework}: We introduce a pipeline that demonstrates the application of self-supervised objectives to multi\hyp{}modal and multi\hyp{}dimensional medical imaging. MultiMedVision serves as an adaptable framework rather than just a single model, enabling the joint training of 2D and 3D data streams.
    \item \textbf{Cross-Dimensional Shared Features}: We demonstrate that a model trained using this framework effectively learns shared features among different dimensions. Our results show that the model learns a modality\hyp{}agnostic latent space, enabling stable performance on both 2D and 3D tasks using a single shared encoder.
\end{enumerate}


\section{Method}

\subsection{SparseMedicalViT: Unified Sparse 2D/3D Architecture}

\subsubsection{Unified Input Processing.}
Given a dataset of mixed 2D and 3D medical images, each 2D image is promoted to a pseudo-volume by zero-padding along the depth axis to match the volumetric patch size $P=14$. Given an input volume $\mathbf{x} \in \mathbb{R}^{C \times Z \times H \times W}$, we partition it into non-overlapping cubic patches $\{\mathbf{p_k}\}$ of size $P^3$, which are linearly projected into $d$-dimensional token embeddings, yielding the sequence $\mathbf{z} = \mathrm{Linear}(\{\mathbf{p_k}\}) \in \mathbb{R}^{S \times d}$, where $S$ denotes the variable-length number of patch tokens. To encode spatial information across all three axes, 3D RoPE is applied to the query and key vectors within each attention layer. For a given grid shape $(G_Z, G_H, G_W)$, we normalize the spatial coordinates along each axis independently to a common scale $[0,\,\alpha]^3$, where $\alpha$ is a position-scale constant~\citep{Dehghani2023}. Specifically, for each axis with grid size $g \in \{G_Z, G_H, G_W\}$, the coordinate of the $i$-th patch along that axis is:
$\mathrm{coord}(i) = \tfrac{\alpha \cdot i}{\max(g - 1,\; 1)}, \; i = 0, \ldots, g - 1.$

This normalization strategy provides two key advantages: (1) it maintains consistent positional encoding scales across varying resolutions and aspect ratios, and (2) it seamlessly handles 2D images ($Z=1$) as a special case where the depth coordinate remains zero. The RoPE frequencies are then computed by allocating equal frequency dimensions to each spatial axis, enabling the model to learn unified representations across heterogeneous 2D and 3D medical imaging modalities.

\subsubsection{Variable-Sized Views, Sequence Packing, and Efficiency.}
Unlike standard approaches that force inputs into fixed resolutions, we adopt a flexible aspect-ratio preserving pipeline. Rather than padding images to a fixed size, we pack patches from all samples into a single continuous sequence, tracking sample boundaries. Attention is then computed directly on this packed sequence using variable-length FlashAttention \citep{dao2022flashattention, dao2023flashattention2}, which uses these boundaries to delineate individual samples, eliminating padding overhead and significantly enhancing training throughput and scalability.

\subsection{Self-Supervised Learning Objective (LeJEPA)}
The core objective learns invariant representations by minimizing the distance between augmented crops (views) of the same input. Let $f_\theta$ be the vision encoder. For each input $\mathbf{x}$, we generate $V_g$ global views that capture broad spatial context and $V_l$ local views that focus on fine-grained details. We compute the global centroid
$\boldsymbol{\mu} = \frac{1}{V_g} \sum_{v=1}^{V_g} f_\theta(\mathbf{x}_v)$
and minimize the prediction loss
$\mathcal{L}_{\mathrm{pred}} = \frac{1}{V} \sum_{v'=1}^{V}
\| \boldsymbol{\mu} - f_\theta(\mathbf{x}_{v'}) \|_2^2$,
where $V = V_g + V_l$ is the total number of views.

To prevent representational collapse, we employ SIGReg~\citep{Balestriero2025}, which enforces the embedding distribution to be approximately isotropic Gaussian, which has been proven to minimize worst-case downstream risk. SIGReg works by comparing the empirical characteristic functions of random 1D projections against the theoretical Gaussian via the Epps--Pulley statistic. The total loss is
$\mathcal{L} = (1 - \lambda)\,\mathcal{L}_{\mathrm{pred}} + \lambda\,\mathcal{L}_{\mathrm{SIGReg}}$.

\subsection{Training Stages}
We investigate the stability and efficacy of joint representation learning through three training protocols, leveraging a pipeline that can ingest a mixed stream of data by concatenating 2D and 3D batches on-the-fly. \textbf{Stage 1 (2D-only)} establishes a baseline by pre-training exclusively on a 2D dataset. \textbf{Stage 2 (Curriculum)} initializes from these 2D weights and continues training with the mixed 2D/3D stream, ensuring continuous exposure to both modalities to prevent catastrophic forgetting and encourage feature alignment. Finally, \textbf{Stage 3 (Native Joint Training)} trains the unified architecture from scratch on the mixed dataset.

\subsection{Experimental Setup}

\subsubsection{Datasets.}
We validate our framework on public datasets. For 2D radiographs, we use the MIMIC-CXR dataset (one frontal image per study) for training (n=133,749) and internal validation (n=33,537), and CheXpert (n=668, radiologist\hyp{}verified) for external testing. For 3D CT, we employ CT-RATE for pre-training (n=46,394) and internal testing (n=3,002), with Rad-ChestCT (n=3,630) serving as an external test set.

\textbf{Implementation Details.}
Images are downsized along the long side (224 for 2D, 112 for 3D) before extracting variable-sized views, as this proof-of-concept study prioritizes computational efficiency. Intensities are normalized to $[-1, 1]$ using min-max scaling for X-rays and custom windows (W=2500, L=250) for CT scans. We extract $V_g=2$ large global views (scale $[0.3, 1.0]$) for semantic context and $V_l=8$ small local views (scale $[0.05, 0.3]$) for fine-grained detail, applied consistently across 2D and 3D inputs. We used a RoPE base of 10,000, with a normalization constant ($\alpha$) of 128. Models are implemented in PyTorch Lightning and trained on 2 NVIDIA RTX PRO 6000 GPUs with BF16 mixed precision. We use the AdamW optimizer (lr=$1.0 \times 10^{-4}$, cosine decay). Stage 1 uses a batch size of 256 (2D). Joint training (Stage 2 \& 3) employs a mixed batch of 192 (2D) and 64 (3D) to respect the proportion of 2D and 3D data. For regularization, we apply weight decay (0.05) and SIGReg loss ($\lambda$=0.025).


\section{Results}

Our evaluation focuses on the intersection of five common labels in both 2D and 3D datasets (Atelectasis, Cardiomegaly, Consolidation, Lung Opacity, and Pleural Effusion). We evaluate our model's performances using linear probing (LP) on the frozen features. Critically, the LP training data uses the same pre-training data, ensuring that neither the encoder nor the LP sees the test set during training.

\subsection{Performance Comparison}

We benchmark our unified model against specialized 2D (BiomedCLIP \citep{Zhang2025}, MedSigLIP \citep{Sellergren2025}, RAD-DINO \citep{Prez-Garca2025}) and 3D (CT-CLIP \citep{Hamamci2026}, AnyMC3D \citep{Liu2025}), as well as supervised baselines trained with the same backbone as SparseMedicalViT.

\begin{table*}[t]
\begin{threeparttable}
\caption{Performance comparison of MultiMedVision (Stage 3) against other 2D and 3D foundation models and supervised baselines.}
\label{tab:tab1}
\fontsize{8pt}{8.2pt}\selectfont
\begin{tabular*}{\textwidth}{@{\extracolsep{\fill}}lccccccc@{}}
\toprule
\multirow{2}{*}{} & \multirow{2}{*}{\begin{tabular}[c]{@{}c@{}}\textbf{LP} \\ \textbf{Training} \\ \textbf{Data} \end{tabular}} & \multicolumn{6}{c}{\textbf{AUROC}}                           \\ \cmidrule(l){3-8}
                  &                                                       & \textbf{ATEL} & \textbf{CARD} & \textbf{CONS} & \textbf{LO}  & \textbf{EFF}  & \textbf{Macro(95\%CI)}           \\ \midrule
\multicolumn{8}{c}{\cellcolor[HTML]{EFEFEF}\textbf{MIMIC (Internal Test Set)}}                                                                                                                     \\
BiomedCLIP(n=15M)\textsuperscript{1}        & 2D                                                                           & 0.78 & 0.8  & 0.82 & 0.75 & 0.88 & 0.81 (0.80-0.81) \\
MedSigLIP(n=33M)\textsuperscript{1}         & 2D                                                                           & 0.83 & 0.83 & 0.85 & 0.79 & 0.91 & 0.84 (0.84-0.85) \\
RAD-DINO(n=0.84M)\textsuperscript{1}          & 2D                                                                           & 0.82 & 0.82 & 0.83 & 0.78 & 0.91 & 0.83 (0.83-0.84) \\
MultiMedVision-S3            & ALL                                                                          & 0.81 & 0.82 & 0.81 & 0.76 & 0.9  & 0.82 (0.81-0.83) \\
Supervised        & ALL                                                                          & 0.78 & 0.79 & 0.77 & 0.71 & 0.85 & 0.78 (0.77-0.79) \\
\multicolumn{8}{c}{\cellcolor[HTML]{EFEFEF}\textbf{CheXpert (External Test Set)}}                                                                                                                  \\
BiomedCLIP\textsuperscript{1}        & 2D                                                                           & 0.82 & 0.86 & 0.9  & 0.86 & 0.9  & 0.87 (0.82-0.90) \\
MedSigLIP\textsuperscript{1}         & 2D                                                                           & 0.88 & 0.89 & 0.9  & 0.9  & 0.94 & 0.90 (0.87-0.93) \\
RAD-DINO\textsuperscript{1}          & 2D                                                                           & 0.87 & 0.87 & 0.89 & 0.85 & 0.92 & 0.88 (0.84-0.91) \\
MultiMedVision-S3            & ALL                                                                          & 0.83 & 0.85 & 0.79 & 0.83 & 0.89 & 0.84 (0.79-0.88) \\
Supervised        & ALL                                                                          & 0.77 & 0.82 & 0.83 & 0.8  & 0.83 & 0.81 (0.76-0.86) \\
\multicolumn{8}{c}{\cellcolor[HTML]{EFEFEF}\textbf{CT-RATE (Internal Test Set)}}                                                                                                                   \\
CT-CLIP(n=46,394)\textsuperscript{2}           & 3D                                                                           & 0.68 & 0.91 & 0.73 & 0.68 & 0.91 & 0.78  \\
AnyMC3D\textsuperscript{3}           & -                                                                     & 0.83 & 0.95 & 0.92 & 0.82 & 0.99 & 0.90  \\
MultiMedVision-S3            & ALL                                                                          & 0.75 & 0.95 & 0.83 & 0.76 & 0.95 & 0.85 (0.83-0.87) \\
Supervised        & ALL                                                                          & 0.72 & 0.93 & 0.75 & 0.66 & 0.94 & 0.80 (0.78-0.82) \\
\multicolumn{8}{c}{\cellcolor[HTML]{EFEFEF}\textbf{RAD-ChestCT (External Test Set)}}                                                                                                               \\
CT-CLIP\textsuperscript{2}             & 3D                                                                           & 0.63 & 0.81 & 0.68 & 0.61 & 0.83 & 0.71  \\
MultiMedVision-S3            & ALL                                                                          & 0.56 & 0.78 & 0.6  & 0.58 & 0.78 & 0.66 (0.64-0.68) \\
Supervised        & ALL                                                                          & 0.56 & 0.73 & 0.6  & 0.55 & 0.76 & 0.64 (0.62-0.66) \\ \bottomrule
\end{tabular*}
\begin{tablenotes}
\scriptsize
  \item LP=Linear Probing, ATEL=Atelectasis, CARD=Cardiomegaly, CONS=Consolidation, LO=Lung Opacity, EFF=Pleural Effusion
  \item[1] Results obtained by training LP on the frozen embeddings of MIMIC training set.
  \item[2] Results reported directly from the original paper; training and test splits are consistent with those used in our experiments.
  \item[3] Results reported directly from the original paper; training data and downstream classifier are \textit{different} from those used in our experiments.
\end{tablenotes}
\end{threeparttable}
\end{table*}

\begin{table*}[t]
\centering
\begin{threeparttable}
\caption{Domain adaptation performance on the external RAD-ChestCT dataset. We compare the AUROC of our model after adapting the LP to the target domain against the pre-adaptation baseline and CT-CLIP.}
\label{tab:tab2}
\fontsize{8pt}{8.2pt}\selectfont
\begin{tabular*}{\textwidth}{@{\extracolsep{\fill}}lccc@{}}
    \toprule
     & \textbf{AUROC(95\%CI)} & \textbf{Improvement} (\%) & \textbf{AUROC (CT-CLIP)} \\
    \midrule
    CARD & 0.83 (0.80-0.87) & 4.94 & 0.81 \\
    ATEL & 0.68 (0.65-0.71) & 12.17 & 0.63 \\
    LO & 0.67 (0.64-0.70) & 8.63 & 0.61 \\
    EFF & 0.86 (0.84-0.88) & 7.87 & 0.83 \\
    CONS & 0.72 (0.68-0.75) & 12.0 & 0.68 \\
    Macro & 0.75 (0.72-0.78) & 9.12 & 0.71 \\
    \bottomrule
\end{tabular*}
\begin{tablenotes}
  \scriptsize
  \item Note: ATEL=Atelectasis, CARD=Cardiomegaly, CONS=Consolidation, LO=Lung Opacity, EFF=Pleural Effusion
\end{tablenotes}
\end{threeparttable}
\end{table*}

Our analysis yields several key findings (\Cref{tab:tab1}). First, MultiMedVision-S3 achieves competitive performance on 2D tasks with a Macro AUROC of 0.82 on MIMIC (comparable to BiomedCLIP's 0.81) and 0.84 on CheXpert (comparable to BiomedCLIP's 0.87). Second, it significantly outperforms CT-CLIP on the internal test dataset (CT-RATE), achieving a Macro AUROC of 0.85 versus CT-CLIP's 0.78. However, on the external RAD-ChestCT dataset, our model underperforms (0.66) compared to CT-CLIP (0.71), suggesting potential domain shift sensitivities. Overall, these results demonstrate a true ``one model, all modalities'' latent space where a single encoder effectively processes both 2D and 3D inputs.

We further analyze domain adapted performance to understand the generalization gap upon 3D external evaluation (\Cref{tab:tab2}). To evaluate this, we train the LP on the RAD-ChestCT training set and evaluate on its test set. This adaptation yields a significant performance improvement (e.g., Macro AUROC increasing from 0.66 to 0.75), suggesting potential factors such as domain shift, resolution differences, and variations in reference standard generation between the internal and external datasets that affect generalization.

\subsection{Training Strategy Analysis}

\begin{table*}[t]
\begin{threeparttable}
\caption{\textbf{Training Strategy Analysis (Left)}. Joint training stages (Stage2, Stage3) significantly improve 3D performance while maintaining competitive 2D performance. \textbf{Modality Robustness Analysis (Right)}. An LP trained on the `ALL' modality strategy effectively leverages shared representations to perform robustly across both 2D and 3D tasks.}
\label{tab:tab3}
\fontsize{8pt}{8.2pt}\selectfont
\begin{tabular*}{\textwidth}{@{\extracolsep{\fill}}ccccccc@{}}
\toprule
\multicolumn{3}{c}{\textbf{Training Strategy Analysis}}                                                                                                  &  & \multicolumn{3}{c}{\textbf{Modality Robustness Analysis}}                                                                                                \\ \cmidrule(r){1-3} \cmidrule(l){5-7}
\textbf{Stage} & \textbf{\begin{tabular}[c]{@{}c@{}}LP\\ Training Data\end{tabular}} & \textbf{\begin{tabular}[c]{@{}c@{}}Macro\\ (95\%CI)\end{tabular}} &  & \textbf{Stage} & \textbf{\begin{tabular}[c]{@{}c@{}}LP\\ Training Data\end{tabular}} & \textbf{\begin{tabular}[c]{@{}c@{}}Macro\\ (95\%CI)\end{tabular}} \\ \cmidrule(r){1-3} \cmidrule(l){5-7}
\multicolumn{3}{c}{\cellcolor[HTML]{EFEFEF}\textbf{MIMIC}}                                                                                               &  & \multicolumn{3}{c}{\cellcolor[HTML]{EFEFEF}\textbf{MIMIC}}                                                                                               \\
1              & ALL                                                                 & 0.82 (0.81-0.83)                                                  &  & 3              & 2D                                                                  & 0.82 (0.82-0.83)                                                  \\
2              &                                                                     & 0.82 (0.81-0.82)                                                  &  &                & 3D                                                                  & 0.53 (0.52-0.54)                                                  \\
3              &                                                                     & 0.82 (0.81-0.83)                                                  &  &                & ALL                                                                 & 0.82 (0.81-0.83)                                                  \\
\multicolumn{3}{c}{\cellcolor[HTML]{EFEFEF}\textbf{CheXpert}}                                                                                            &  & \multicolumn{3}{c}{\cellcolor[HTML]{EFEFEF}\textbf{CheXpert}}                                                                                            \\
1              & ALL                                                                 & 0.84 (0.80-0.88)                                                  &  & 3              & 2D                                                                  & 0.84 (0.80-0.88)                                                  \\
2              &                                                                     & 0.85 (0.80-0.88)                                                  &  &                & 3D                                                                  & 0.55 (0.48-0.61)                                                  \\
3              &                                                                     & 0.84 (0.79-0.88)                                                  &  &                & ALL                                                                 & 0.84 (0.79-0.88)                                                  \\
\multicolumn{3}{c}{\cellcolor[HTML]{EFEFEF}\textbf{CT-RATE}}                                                                                             &  & \multicolumn{3}{c}{\cellcolor[HTML]{EFEFEF}\textbf{CT-RATE}}                                                                                             \\
1              & ALL                                                                 & 0.74 (0.72-0.76)                                                  &  & 3              & 2D                                                                  & 0.58 (0.56-0.61)                                                  \\
2              &                                                                     & 0.83 (0.81-0.84)                                                  &  &                & 3D                                                                  & 0.86 (0.84-0.87)                                                  \\
3              &                                                                     & 0.85 (0.83-0.87)                                                  &  &                & ALL                                                                 & 0.85 (0.83-0.87)                                                  \\
\multicolumn{3}{c}{\cellcolor[HTML]{EFEFEF}\textbf{RAD-ChestCT}}                                                                                         &  & \multicolumn{3}{c}{\cellcolor[HTML]{EFEFEF}\textbf{RAD-ChestCT}}                                                                                         \\
1              & ALL                                                                 & 0.59 (0.57-0.61)                                                  &  & 3              & 2D                                                                  & 0.49 (0.47-0.52)                                                  \\
2              &                                                                     & 0.60 (0.58-0.62)                                                  &  &                & 3D                                                                  & 0.65 (0.63-0.67)                                                  \\
3              &                                                                     & 0.66 (0.64-0.68)                                                  &  &                & ALL                                                                 & 0.66 (0.64-0.68)                                                  \\ \bottomrule
\end{tabular*}
\end{threeparttable}
\end{table*}

We analyze the stability and efficacy of joint representation learning by comparing Stage 1 (2D-only), Stage 2 (Curriculum), and Stage 3 (Native Joint). The LP is trained on ``ALL'' inputs (both 2D and 3D pre-training data) (\Cref{tab:tab3}).

The addition of 3D data in Stages 2 and 3 improves 3D performance compared to the 2D-only baseline. On CT-RATE, Macro AUROC jumps from 0.74 (Stage 1) to 0.83 (Stage 2) and 0.85 (Stage 3), confirming that robust 3D representations require native volumetric exposure. This trend also appears in external evaluation on RAD-ChestCT, where both joint strategies outperform the baseline, with Stage 3 reaching 0.66 and Stage 2 reaching 0.60. Moreover, 2D performance remains stable across all stages: MIMIC stays around 0.82 and CheXpert fluctuates slightly (0.84 to 0.85), indicating that the addition of 3D data injects valuable geometric context without interfering with 2D feature learning. Lastly, Stage 2 (Curriculum) achieves competitive performance with Stage 3 (Native Joint) across tasks, highlighting the flexibility of the framework: a model originally trained on 2D images can be effectively adapted for both 2D and 3D tasks.

\subsection{Modality Robustness Analysis}

We evaluate feature redundancy and specificity of the vision encoder trained on both 2D and 3D (Stage 3) by training the LP on specific modalities (2D and 3D, or ALL) and evaluating on both (\Cref{tab:tab3}).

Our findings reveal a dual nature of the learned representations: there are modality-specific features, evidenced by the poor cross-modality performance of single-modality probes (e.g., 2D probe failing on CT-RATE (3D), 0.58; 3D probe failing on MIMIC (2D), 0.53). Crucially, the LP trained on ``ALL'' inputs effectively leverages these shared features to perform robustly across modalities. On CT-RATE (3D), the ``ALL'' probe achieves 0.85 Macro AUROC, matching the specific 3D-trained probe (0.86). Similarly, on MIMIC (2D), the ``ALL'' probe maintains high performance (0.82), identical to the 2D-specific probe. This confirms that the joint training strategy fosters a unified, modality-agnostic latent space capable of generalizing across dimensionalities. Similar trends are observed in CheXpert and RAD-ChestCT.

\subsection{Qualitative Results}

\begin{figure*}[t]
    \centering
    \includegraphics[width=0.75\linewidth]{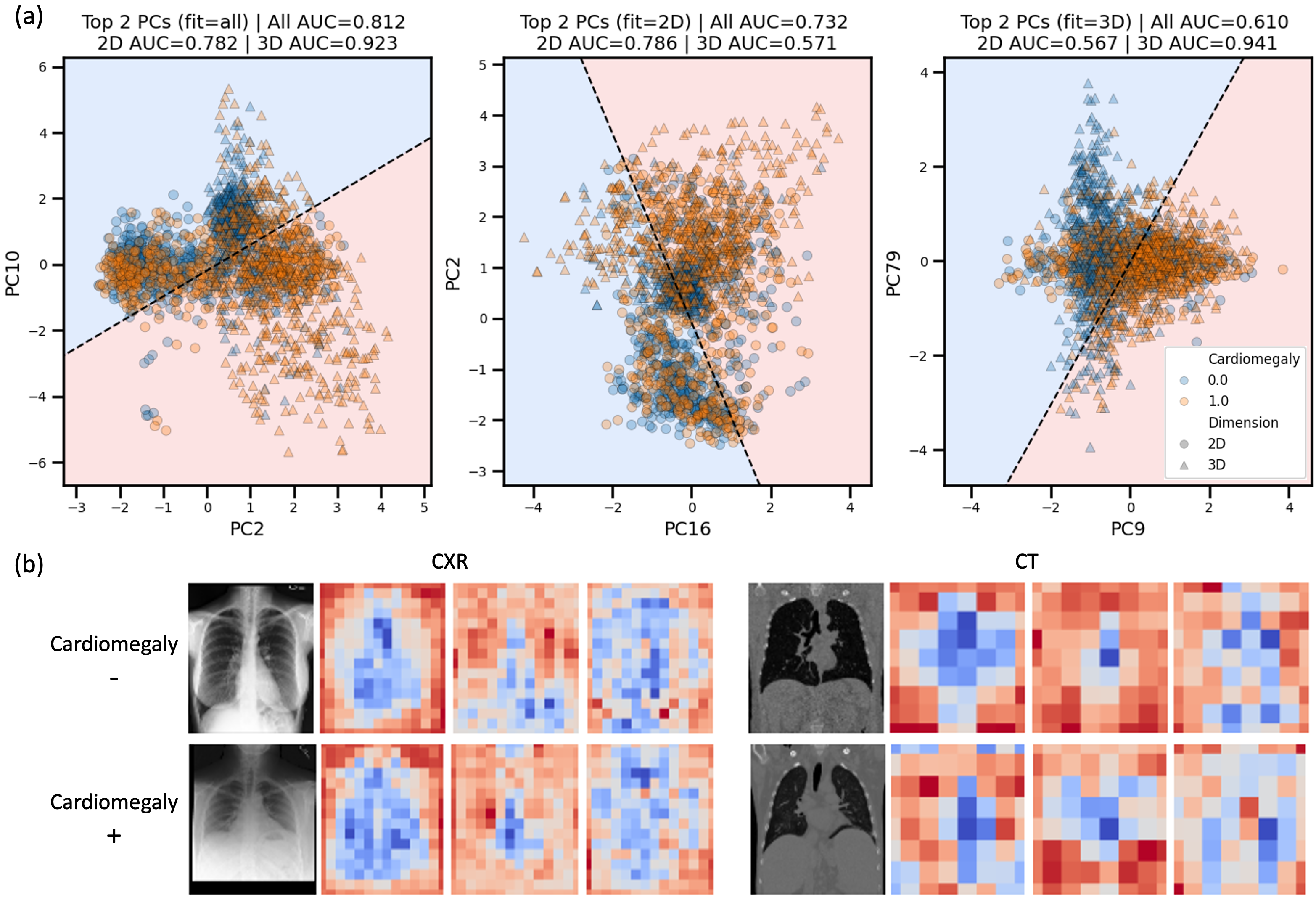}
    \caption{(a) Logistic regression on PCA-projected features reveals that modality-specific models fail to generalize across dimensions: the 2D-fitted model cannot predict Cardiomegaly in CT, and the 3D-fitted model cannot predict it in CXR. In contrast, a model fit on joint 2D/3D features maintains competitive performance across both modalities, leveraging shared principal components. Notably, each model relies on distinct PCs for prediction, suggesting modality-specific and shared feature subspaces coexist within the unified representation. (b) DINO-style PCA feature maps reveal spatially similar activation patterns associated with Cardiomegaly across both CXR and CT modalities, indicating that the encoder captures cross-modal anatomical correspondence.}
    \label{fig:Fig2}
\end{figure*}

Principal Component Analysis (PCA) of the embeddings reveals that while some dimensions are modality-specific (e.g., predicting Cardiomegaly primarily from 2D or 3D features), there are common principal components that are predictive across both modalities (\cref{fig:Fig2}a). This supports the Modality Robustness finding (Section~3.3) that the model learns both shared and specific representations in a unified space. We visualize discriminative features using DINO-style PCA maps \citep{Oquab2024} (\cref{fig:Fig2}b), which highlight regions relevant to specific pathologies (e.g.\ Cardiomegaly). These visualizations suggest that the model learns semantically meaningful similar features across modalities.

\section{Conclusion}

We present MultiMedVision, a unified framework that enables a single shared encoder to learn from both 2D radiographs and 3D volumetric scans through unified positional encodings and variable-length sequence packing, without requiring modality-specific adapters. Our results show competitive 2D and 3D performance over existing baselines, while analysis of the learned representations reveals both modality-specific and shared features that enable cross-modality prediction. A key limitation is that computational constraints prevented evaluation of scaling laws beyond the current $\sim$200k-image training set. Future work will extend the framework to additional modalities such as MRI and investigate the scaling behavior of the SparseMedicalViT architecture on larger, more diverse datasets.

\bibliography{references}

\end{document}